\documentclass{article}
\usepackage{spconf,amsmath,graphicx}
\usepackage{url}

\def\Fig#1{{Fig.\ \ref{fig:#1}}}
\def\Eq#1{{Eq.\ \ref{eq:#1}}}

\let\OldCaption=\caption
\renewcommand{\caption}[1]{\small\OldCaption{\em#1}}

\makeatletter
\def\@normalsize{\@setsize\normalsize{10pt}\xpt\@xpt
\abovedisplayskip 10pt plus2pt minus5pt\belowdisplayskip
\abovedisplayskip \abovedisplayshortskip \z@
plus3pt\belowdisplayshortskip 6pt plus3pt
minus3pt\let\@listi\@listI}
\def\subsize{\@setsize\subsize{12pt}\xipt\@xipt}
\def\section{\@startsection {section}{1}{\z@}{1.0ex plus
1ex minus .2ex}{.2ex plus .2ex}{\large\bf}}
\def\subsection{\@startsection {subsection}{2}{\z@}{.2ex
plus 1ex} {.2ex plus .2ex}{\subsize\bf}} \makeatother

\def\@listI{%
 \leftmargin\leftmargini
 \partopsep 0pt
 \parsep 0pt
 \topsep 0pt
 \itemsep pt
 \relax
} \long\def\@makecaption#1#2{
 \vskip -5pt
 \setbox\@tempboxa\hbox{\small{#1\,:\,#2}}
  \ifdim \wd\@tempboxa >\hsize \unhbox\@tempboxa\par \else
  \hbox to\hsize{\hfil\box\@tempboxa\hfil}
\fi \vskip -0.2cm}

\title{Unsupervised Body Part Regression via Spatially Self-ordering Convolutional Neural Networks}
\name{Ke Yan ~~~~~~Le Lu ~~~~~~Ronald M. Summers}
\address{Imaging Biomarkers and Computer-Aided Diagnosis Lab, Clinical Image Processing Service \\ Radiology and Imaging Sciences, National Institutes of Health Clinical Center}
\begin{document}
%
\maketitle
\begin{abstract}

Automatic body part recognition for CT slices can benefit various medical image applications. Recent deep learning methods demonstrate promising performance, with the requirement of large amounts of labeled images for training. The intrinsic structural or superior-inferior slice ordering information in CT volumes is not fully exploited. In this paper, we propose a convolutional neural network (CNN) based Unsupervised Body part Regression (UBR) algorithm to address this problem. A novel unsupervised learning method and two inter-sample CNN loss functions are presented. Distinct from previous work, UBR builds a coordinate system for the human body and outputs a continuous score for each axial slice, representing the normalized position of the body part in the slice. The training process of UBR resembles a self-organization process: slice scores are learned from inter-slice relationships. The training samples are unlabeled CT volumes that are abundant, thus no extra annotation effort is needed. UBR is simple, fast, and accurate. Quantitative and qualitative experiments validate its effectiveness. In addition, we show two applications of UBR in network initialization and anomaly detection.
\end{abstract}
\begin{keywords}
Body Part Recognition, Unsupervised Learning, Convolutional Neural Network, Slice Ordering
\end{keywords}
\section{Introduction}
\label{sec:intro}

Body part recognition is ubiquitously useful in medical image applications, such as automatic scan range planning and providing body spatial priors to initialize computer aided detection (CADe) and diagnosis (CADx) systems, and so on \cite{yan2016multi}. Traditional methods normally use hand-crafted image classification features \cite{yan2016multi}. Lately, deep learning approaches \cite{yan2016multi, roth2015anatomy, Zhang2017SelfSup} have been adopted with promising results where Convolutional neural networks (CNN) are employed to learn deep image features. However, large amounts of manually labeled training image data are required \cite{yan2016multi, roth2015anatomy, Zhang2017SelfSup}. Recently, Zhang et al.\ \cite{Zhang2017SelfSup} propose a self-supervised method that permits pre-training a CNN model in an unsupervised manner. Nevertheless, it still needs fine-grained labeled CT slices to supervisedly fine-tune the network, before predicting labels of anatomical body parts.

In this paper, we present an Unsupervised Body part Regression or Regressor (UBR) that entirely learns from unlabeled CT volumes. The training volumes can have any scan range (chest, abdomen, pelvis, etc.), hence are abundant in every hospital's picture archiving and communication system (PACS). The superior-inferior slice ordering information is leveraged to train UBR. This intuition somewhat resembles the unsupervised pre-training scheme \cite{Zhang2017SelfSup} but our training procedure, network structure, and loss function are all different. UBR is also more efficient, accurate from our empirical evaluation and requires no labeled CT images completely. By minimizing an order loss and a distance loss, UBR learns the body part knowledge from inter-slice relationships in a self-organization process (defined as ``some overall order arises from local interactions between parts of an initially disordered system'' \cite{wikiSO}).

Most previous work splits the whole body into several discrete anatomical parts \cite{yan2016multi, roth2015anatomy}, which may be hard to be precisely defined (problematic at transition regions \cite{roth2015anatomy}) and have limitations or constraints on different applications. More importantly, they cannot discriminate slices inside a part. In contrast, UBR is a continuous-valued regressor. It builds an axial coordinate system for the body and outputs a continuous score for each slice, which represents the normalized position of the body part in the slice. Thus, it is fine-grained and useful for identifying different parts of the body. Experimental results show that it outperforms \cite{Zhang2017SelfSup} where 88 extra labeled CT volumes are required to fine-tune. Besides body part recognition, we demonstrate two other applications of UBR: CNN weight transfer learning in a CADe task; and detection of significant anomalies in CT volumes (e.g., scan artifacts or large lesions). 

\section{Method}
\label{sec:method}

\begin{figure}[htb]
	\begin{minipage}[b]{1.0\linewidth}
		\centering
		\centerline{\includegraphics[width=8.5cm,trim=120 90 230 120,clip]{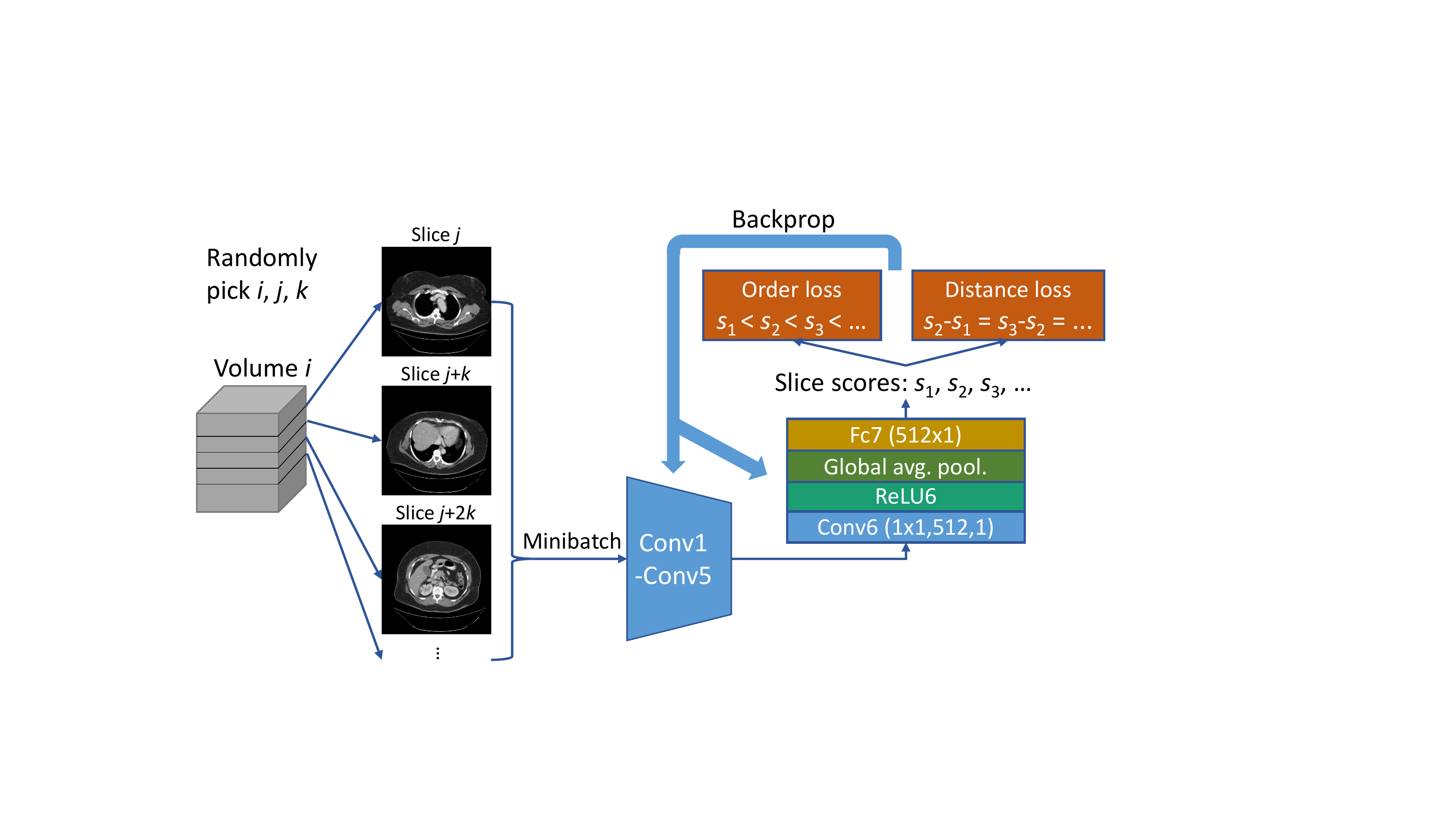}}
	\end{minipage}
	\caption{Framework of the proposed unsupervised body part regression (UBR).}
	\label{fig:framework}
\end{figure}

{\bf Motivation:} Volumetric medical images are intrinsically structured where the position and appearance of organs are relatively aligned. Our idea is to predict a continuous score for each axial CT slice as the normalized body coordinate value. As the image slice index in a volume increases (in the superior-inferior order), the predicted body coordinate scores should become larger accordingly. Note that the image volumes extracted from PACS often have different scan ranges (e.g., start, end, and inter-slice intervals), hence the slice indices cannot be directly used as slice scores or as labels to learn the scores. In this paper, we enforce the deeply learned regressor to obey the spatial superior-inferior ordering as a hard constraint. In addition, any numeric difference of the UBR predicted slice scores should be approximately proportional to the spatial distance between slice indices. 
Following this intuition, we propose the unsupervised body part regressor (UBR), see \Fig{framework}.

{\bf Selection of Training Samples:} In each training iteration, we randomly select $ g $ volumes, then randomly pick $ m $ equidistant slices from each selected volume. A starting slice $ j $ and a slice interval $ k $ are also randomly determined (\Fig{framework}). 

{\bf Network Architecture:} The first few layers of the network include convolution, rectified linear unit (ReLU), and max pooling. The parameters can be directly adopted from ImageNet pre-trained CNN models, e.g., the conv1--conv5 in AlexNet \cite{Krizhevsky2012AlexNet} or VGG-16 \cite{Simonyan2015Vgg}. After these layers, we add a new convolutional layer, Conv6, with 512~1$ \times $1 filters and stride 1, followed by a ReLU layer. Conv1--Conv6 are used to learn discriminative deep image features for body part recognition. Then a global average pooling layer is attached to summarize each of the 512 activation maps to one value, leading to a 512D feature vector. It makes the network structure robust to the position of the body in the slice. At last, a fully connected layer (Fc7) projects the feature vector to the slice score. 

{\bf Loss Function:} The loss function is critical for the proposed unsupervised deep learning method. We learn the ordering relationship between slice scores as a surrogate which can be obtained for free, so annotated anatomical labels \cite{roth2015anatomy,yan2016multi} are not required. A similar idea is to use the Siamese network \cite{bromley1994siamese}, but we have extended it by using more than two samples as a group and adopting two inter-subject loss terms. Experimental results in Section \ref{sec:exp} will demonstrate the superiority of our proposed strategies. 

The first loss term is the order loss $ L_{\rm order} $, which requires slices with larger indices to have larger scores. As expressed in \Eq{orderLoss}, $ L_{\rm order} $ is a logistic loss. $ g $ is the number of CT volumes in a mini-batch; $ m $ is the number of image slices in each volume; $ S(i,j) $ is the slice score of slice $ j $ in volume $ i $; $ h $ is the sigmoid activation function.
\begin{equation} \label{eq:orderLoss}
	L_{\rm order}=-\sum_{i=1}^g{\sum_{j=1}^{m-1}{\log h\left(S(i,j+1)-S(i,j)\right)}}.
\end{equation}

Besides keeping the qualitative order of the regressed slice scores, we also constrain them to increase linearly. The numeric difference between two slice scores should be proportional to the physical distance between the two images. Because we intentionally pick the sets of equidistant slices (e.g., slices $ j, j+k, j+2k, \ldots $), the slice scores should be equidistant as well. The distance loss is thus defined as:
\begin{equation}\label{eq:distLoss}
	\begin{aligned}
	      L_{\rm dist} =& \sum_{i=1}^g{\sum_{j=1}^{m-2}{f(\Delta_{i,j+2}-\Delta_{i,j+1})}}, \\
		\Delta_{i,j} =&\, S(i,j)-S(i,j-1),
	\end{aligned}
\end{equation}
where $ f $ is the smooth L1 loss \cite{Ren2015Faster}. The final loss is
\begin{equation}\label{eq:finalLoss}
	L =  L_{\rm order} + L_{\rm dist}.
\end{equation}

In the training process, the order loss and distance loss terms collaborate to ``push'' each slice score towards the correct direction relative to other slices. If the order loss does not exist, a trivial solution may be obtained where all slice scores are constant. If the distance loss is absent, the slice scores will be nonlinear and less accurate.


\section{Experiments}
\label{sec:exp}

{\bf Datasets and Implementation Details:} We collect 800 random unlabeled CT volumes of 420 subjects from our hospital PACS as our UBR training dataset. Volumes are in the range of 30$\sim$700 slices each. Most volumes are chest-abdomen-pelvis scans and the exact body scan ranges are not used during training. The testing set includes 18,195 CT slices randomly sampled from 260 CT volumes of 140 new subjects. To assess the UBR's quantitative performance, each testing slice is manually labeled as one of the 3 common classes: chest (5903 slices), abdomen (6744), or pelvis (5548). The abdomen class starts from the upper border of the liver and ends at the upper border of the ilium. The proposed method can be easily extended to recognize other parts of the body. The data have various pixel spacings (0.6--1.0 mm), reconstruction kernels, and pathological conditions.

As shown in \Fig{framework}, the layers Conv1--Conv5 of UBR are the same as those in VGG-16 \cite{Simonyan2015Vgg}, initialized using the ImageNet \cite{Feifei09ImageNet} pre-trained model. Conv6 and Fc7 are trained from scratch where the number of volumes per mini-batch is set $g=12$ and the number of sampled slices per volume $m=8$. The UBR network is trained using stochastic gradient descent with the initial learning rate 0.002 and converged in 1.5K iterations, taking only 12 min on a Titan X Pascal GPU. The inference time per slice is 4 ms.

\subsection{Qualitative and Quantitative Results}
\label{ssec:res_detail}

\begin{figure}[]
	\begin{minipage}[b]{1.0\linewidth}
		\centering
		\centerline{\includegraphics[width=8.2cm]{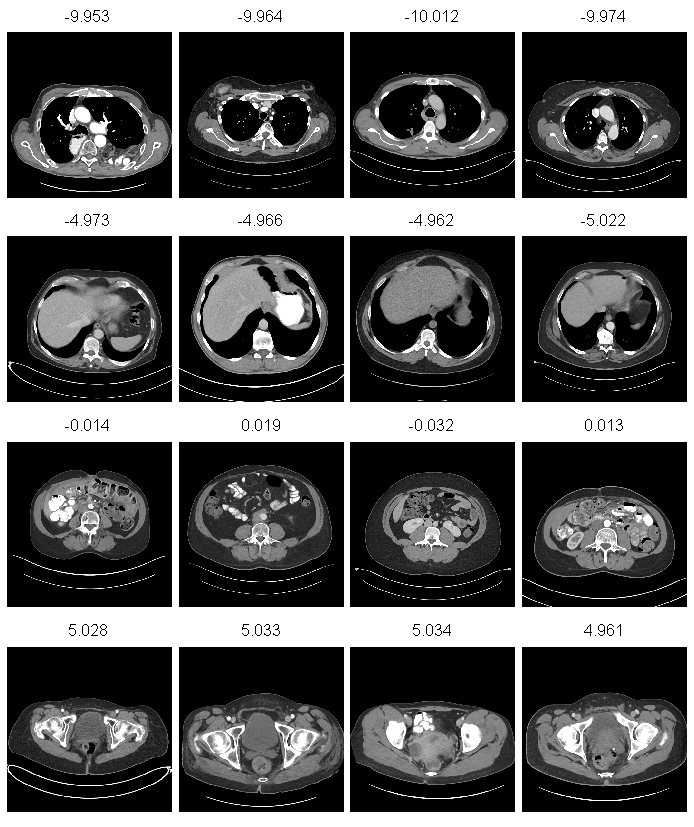}}
	\end{minipage}
	\caption{Sampled slices with slice scores close to -10, -5, 0, and 5, respectively. Images in the same row show similar scores and body parts (randomly picked from the testing set). The numbers above each slice are UBR regressed scores.}
	\label{fig:query_res}
\end{figure}

Under our defined loss function (\Eq{finalLoss}), the UBR output scores mostly range between -15 and 15. \Fig{query_res} illustrates some qualitative results. It can be observed that the deeply learned regression scores and human anatomical body parts correspond well (-10: upper chest, -5: liver dome, 0: lower abdomen, 5: lower pelvis). UBR is also robust to the varying position, size and pathological conditions (e.g., row 1 column 1, atypical presentation of bowel in the chest) of the human body. 

To evaluate the quantitative performance of UBR, we classify the testing slices into three classes only based on the UBR regressed slice scores. Two extra validation CT volumes are used to determine two thresholds to separate three body zones. We also implement two methods for baseline comparison: supervisedly training a 3-class classifier using the labeled slices,  similar to \cite{roth2015anatomy}; and the self-supervised method in \cite{Zhang2017SelfSup}. It first pre-trains the network with unlabeled slice pairs, then fine-tunes the network with fine-grained labeled volumes to learn slice scores. In the pre-training stage, \cite{Zhang2017SelfSup} concatenates the Fc6 features of two slices to predict the order relationship between the slices. It cannot use the multi-slice order loss or distance loss. We calibrate two thresholds for it like UBR. Note that both \cite{roth2015anatomy} and \cite{Zhang2017SelfSup} require labeled samples in training, so we further manually annotated 88 volumes from 88 new subjects to facilitate \cite{roth2015anatomy,Zhang2017SelfSup}. The performance comparison results are displayed in Table \ref{tbl:acc}. 

\begin{table}
	\centering
	\setlength{\tabcolsep}{5pt}
	\renewcommand{\arraystretch}{1.2}
	\begin{tabular}{ll}
		\hline 
		Method & Acc. (\%) \\
		\hline
		Supervised \cite{roth2015anatomy} with 88 labeled volumes	& 98.84 \\
		Self-supervised \cite{Zhang2017SelfSup} with 88 labeled volumes	& 95.28 \\
		Self-supervised \cite{Zhang2017SelfSup} with 2 labeled volumes	& 72.05 \\
		UBR with 2 labeled volumes	& 95.99 \\
		UBR (AlexNet) with 2 labeled volumes	& 95.61 \\
		UBR, pool6 features with 88 labeled volumes	& 98.41 \\
		\hline
	\end{tabular}
	\caption{Accuracy comparison for body part classification.}
	\label{tbl:acc}
\end{table}

The supervised classification similar to the method of \cite{roth2015anatomy} achieves the highest accuracy of 98.84\%. However, this method needs extra 88 labeled volumes for training. In addition, it is dedicated to classification, which is arguably an easier task than the fine-grained regression in \cite{Zhang2017SelfSup} and UBR. Using only 2 labeled volumes to calibrate thresholds on the regression score, UBR outperforms the self-supervised approach \cite{Zhang2017SelfSup} which are fine-tuned with 88 labeled volumes (95.99\% versus 95.28\%). The classification accuracy of fine-tuning the network of \cite{Zhang2017SelfSup} with only two extra labeled volumes (as the same condition of UBR) is noticeably inferior (only at 72.05\%). Finally, we explore using the 512D feature vector from pool6 of UBR to train a logistic regressor on the 88 labeled volumes. An improved accuracy is obtained of 98.41\%. The pool6 features contain information related to body parts, and can be used to train a more complex classifier when labeled samples are presented. 

\begin{figure*}[t]
\centerline{
\begin{tabular}{ccc}
    \includegraphics[width=0.30\linewidth]{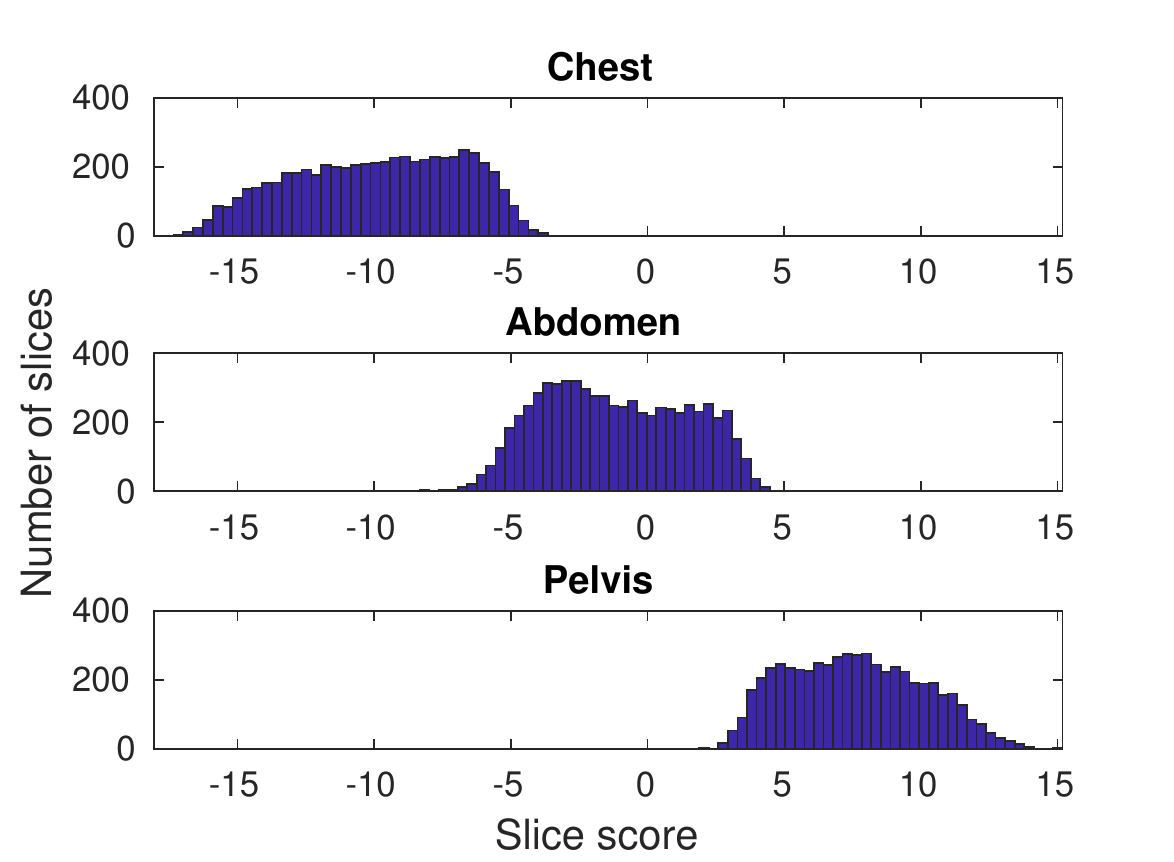} &
    \includegraphics[width=0.30\linewidth]{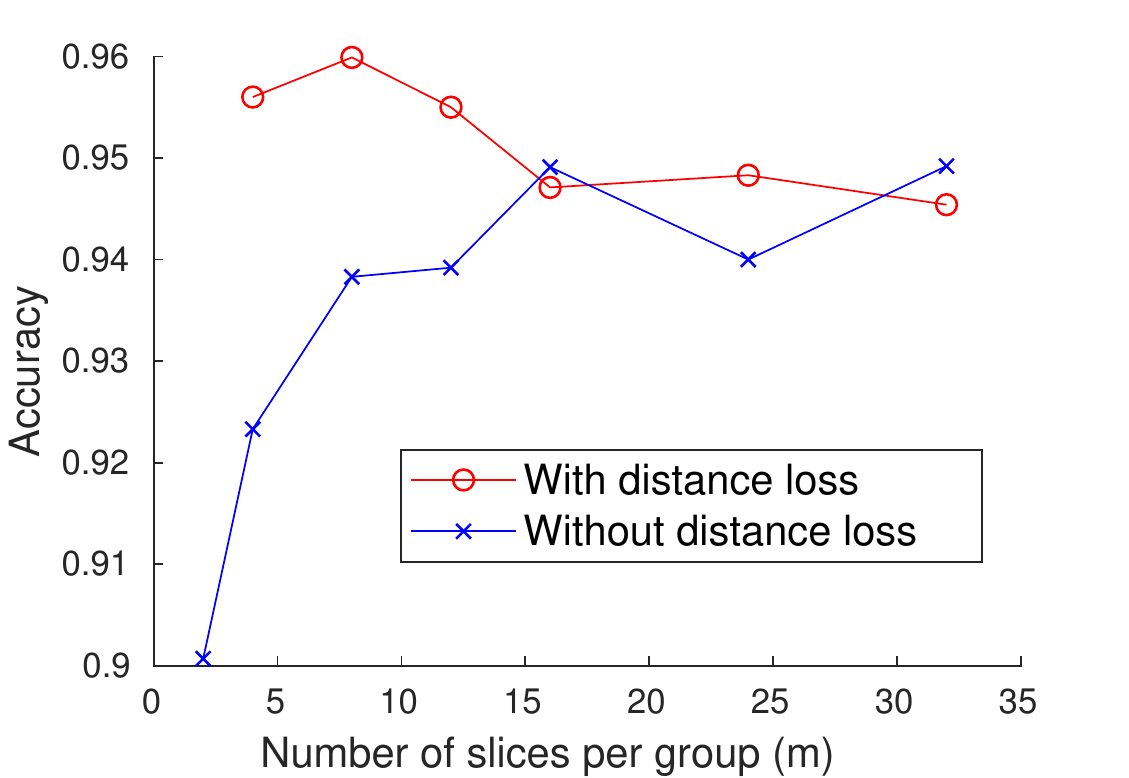} &
		\includegraphics[width=0.30\linewidth]{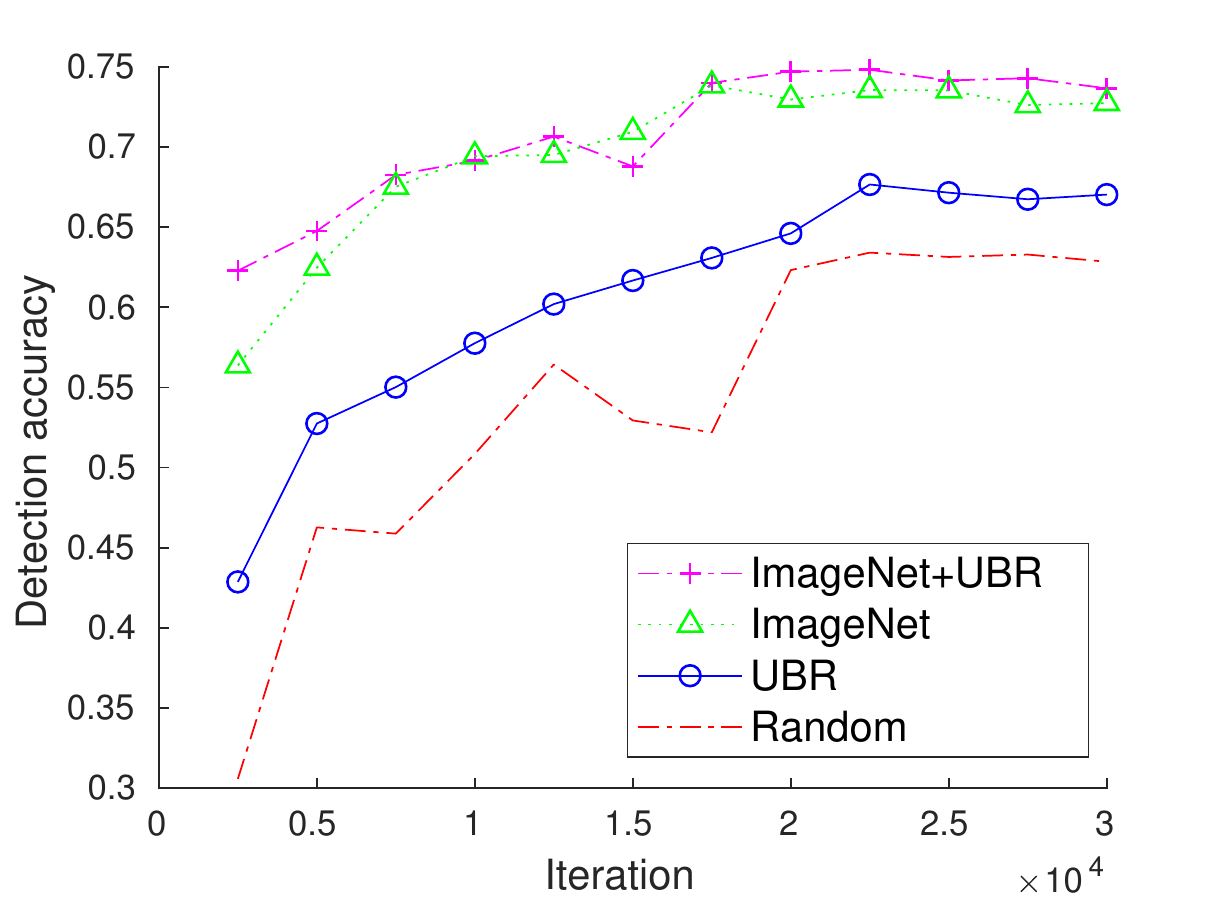} \\
		(a) & (b) & (c)\\
    \end{tabular}
}
\caption{(a) Histogram of slice scores in each class in the test set; 
(b) Accuracy of body part classification of UBR with different parameters; 
(c) Detection accuracy with different initialization methods on the validation set of the DeepLesion dataset \cite{Yan2017DeepLesion}. 
} 
\label{fig:results}
\end{figure*}

To analyze the classification errors when using UBR scores, we draw the histogram of slice scores in each class in \Fig{results} (a). It can be found that most classification errors appear at transition regions because of their intrinsic ambiguity. This will not be a problem in practice since UBR predicts continuous body coordinate scores. The scores for certain body parts may become inaccurate if there are too few slice samples of these parts existing in our UBR (unlabeled) training dataset. 

%
%

Choosing a proper $ m $ can improve the body part recognition accuracy, as indicated by \Fig{results} (b). When increasing $ m $, we also decrease $ g $ to keep $ m\times g=96 $. If $ m=2 $, the distance loss cannot be applied and the order loss is the same with \cite{Zhang2017SelfSup}. Its result is inferior than those with $ m>2 $, proving the effectiveness of the proposed multi-slice strategy. With more CT slices per group, the inter-slice relationship can be better regulated, especially when the distance loss is absent. Incorporating the distance loss (\Eq{finalLoss}) can also increase the accuracy, especially with $m \leq 12$. When there are too many slices per group ($m \geq 16$) to maintain the distance loss, this constraint may become overly strict to be more effective.

\subsection{Applications}
\label{ssec:app_init}
 
{\bf Transfer learning} is very critical for deep learning based medical image problems. Medical datasets are often at small scales versus the huge numbers of parameters in deep neural network models so that training a CNN from scratch can be challenging. \cite{greens2016overview, Shin2016TMI} have found that initializing the network weights adopted from ImageNet \cite{Feifei09ImageNet} pre-trained CNN models is an effective strategy. In this section, we explore the feasibility of using the trained UBR network on the initialization of CNNs for a new CAD task, using  DeepLesion dataset \cite{Yan2017DeepLesion}. It contains 32120 CT slices of size 512$ \times $512 from 10594 studies of 4459 unique patients where bounding-boxes are annotated by radiologists on each slice, marking a variety of lesions or tumors (e.g., lung nodules, lymph nodes, liver/kidney lesions and so on). We adopt the Faster RCNN \cite{Ren2015Faster} method (built upon VGG-16 network) for lesion detection as a binary problem (lesion vs.\ non-lesion). 

Four initialization strategies are exploited for DeepLesion: Random (training a Faster RCNN from scratch), UBR (using a UBR model trained from scratch to initialize Faster RCNN), ImageNet (using a VGG-16 model trained on ImageNet to initialize Faster RCNN), and ImageNet+UBR (first using an ImageNet model to initialize UBR, then using the fine-tuned UBR to initialize Faster RCNN). The learning rates start from 0.002 and are reduced by a factor of 10 in every 20K, 20K, 15K and 15K iterations, respectively. The detection accuracy on DeepLesion is described in \Fig{results} (c). The accuracy is the average recall of top-5 detections on each slice. A predicted box is treated as correct if the intersection-over-union measurement between itself and a ground-truth box is larger than 0.5. The performance of {\bf DeepLesion-UBR} is significantly better than {\bf DeepLesion-Random}. However {\bf DeepLesion-ImageNet} still outperforms {\bf DeepLesion-UBR}, probably because ImageNet \cite{Feifei09ImageNet} is a much extensive and labeled image dataset. The ImageNet pre-trained CNN model has learned very comprehensive sets of filters, especially those representing fine image texture features that are effective for detecting small lesions in CT images. Last, the doubly fine-tuned {\bf DeepLesion-ImageNet+UBR} scheme achieves higher initial accuracies (at 2.5K and 5K iterations) and a slightly improved final accuracy (74.82\% vs.\ 73.84\%), compared to {\bf DeepLesion-ImageNet}. The probable reason is that UBR can further familiarize {\bf DeepLesion-ImageNet} model with the statistics of CT images \cite{Yan2017DeepLesion}, and keep mostly the convolutional filters learned from ImageNet \cite{Feifei09ImageNet}.

\begin{figure}[]
	\begin{minipage}[b]{1.0\linewidth}
		\centering
		\centerline{\includegraphics[width=8.5cm,trim=30 0 40 0,clip]{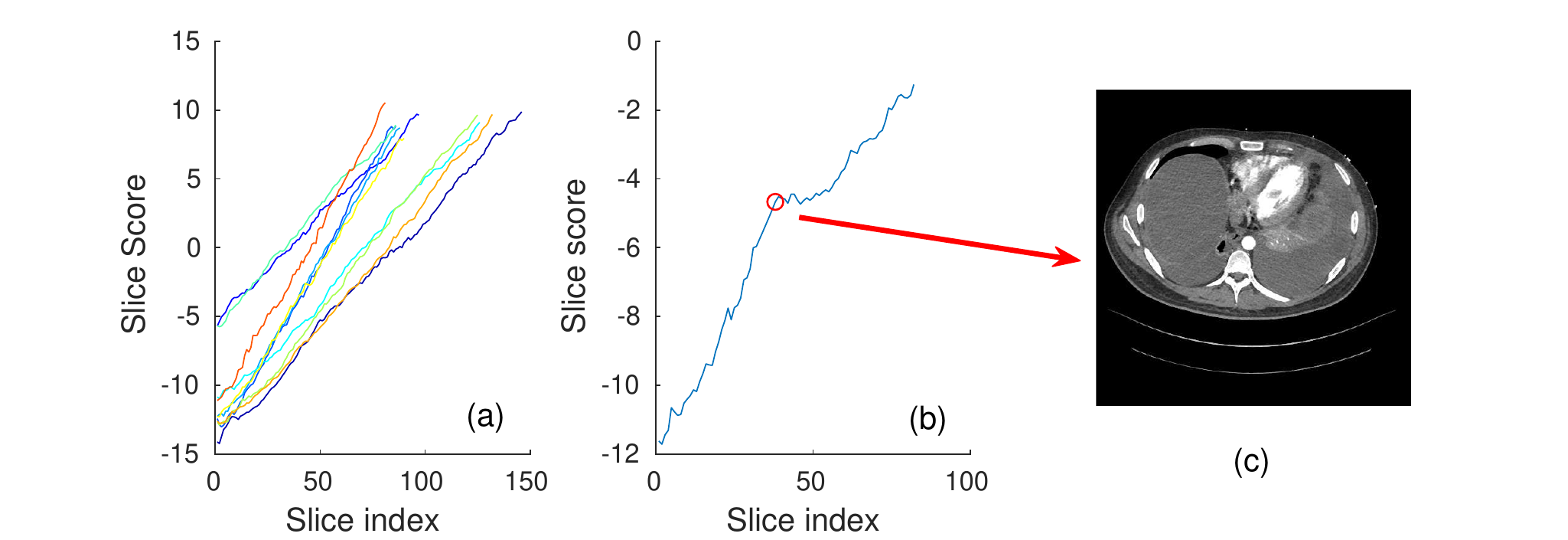}}
	\end{minipage}
	\caption{Examples of anomaly detection using UBR. (a). Slice score curves in 10 normal volumes. (b)--(c). Slice score curve and sample slice in an abnormal CT volume.}
	\label{fig:anomaly}
\end{figure}

{\bf Anomaly Detection:} UBR is able to mine images with significant abnormal appearance (rare in the training set), such as scanning artifacts and large lesions. A simple method is to calculate and plot the body coordinate scores per slice in any CT scan (as $ y $-axis) against the slice indices (as $ x $-axis), to obtain a curve or trajectory. From the curve, we identify several volumes with abnormal non-smooth ``turns''. For the normal volumes (\Fig{anomaly} (a)), their plots are roughly linear with small noise variations indicating inter-subject variances. The initial and final ends of each curve represent the anatomical scan range and the slopes correlate with the slice intervals. We compute the correlation coefficient ($ r $) between the slice indices and slice scores. \Fig{anomaly} (b) is an exemplary abnormal volume with $ r < 0.99$. By manual examination, we find that this nonlinearity is caused by the ascitic fluid appeared in chest and abdomen (\Fig{anomaly} (c)). Besides, 15 volumes in the test set were found to have atypical scan directions (not monotonically from chest to pelvis).

\section{Conclusion}
\label{sec:conclusion}

In this paper, we present an Unsupervised Body part Regressor (UBR) that learns a normalized body coordinate system representing anatomical body parts from unlabeled CT scans. Our method is simple and effective, with zero manual annotation effort needed. Quantitative experimental results and two different applications or extensions have demonstrated its promising performance and good utilities. 


\bibliographystyle{IEEEbib}
\bibliography{refs}

\end{document}